\documentclass[journal]{IEEEtran}
\usepackage{amsmath} % assumes amsmath package installed
\usepackage{amssymb}  % assumes amsmath package installed
\usepackage{hyperref}
\usepackage{multirow}
\usepackage{subcaption}
\usepackage[pdftex]{graphicx}
\usepackage[]{algorithm2e}
\usepackage{lipsum}
\usepackage{enumerate}
\usepackage{tabularx}
\usepackage{color}
\usepackage[english]{babel}
\usepackage{blindtext}

\usepackage{array}
\newcolumntype{L}[1]{>{\raggedright\let\newline\\\arraybackslash\hspace{0pt}}m{#1}}
\newcolumntype{C}[1]{>{\centering\let\newline\\\arraybackslash\hspace{0pt}}m{#1}}
\newcolumntype{R}[1]{>{\raggedleft\let\newline\\\arraybackslash\hspace{0pt}}m{#1}}

\newcommand\T{\rule{0pt}{2.6ex}}       % Top strut
\newcommand\B{\rule[-1.2ex]{0pt}{0pt}} % Bottom strut

\ifCLASSINFOpdf
\else
\fi

\begin{document}
% \title{Bare Demo of IEEEtran.cls\\ for IEEE Journals}

% \author{Michael~Shell,~\IEEEmembership{Member,~IEEE,}
%         John~Doe,~\IEEEmembership{Fellow,~OSA,}
%         and~Jane~Doe,~\IEEEmembership{Life~Fellow,~IEEE}% <-this % stops a space
% \thanks{M. Shell was with the Department
% of Electrical and Computer Engineering, Georgia Institute of Technology, Atlanta,
% GA, 30332 USA e-mail: (see http://www.michaelshell.org/contact.html).}% <-this % stops a space
% \thanks{J. Doe and J. Doe are with Anonymous University.}% <-this % stops a space
% \thanks{Manuscript received April 19, 2005; revised August 26, 2015.}}

\title{An Occluded Stacked Hourglass Approach to Facial Landmark Localization and Occlusion Estimation}
%\title{Coarse Gaze Direction Estimation of Vehicle Occupants using Deep CNN}

\author{Kevan Yuen and Mohan M. Trivedi\\
  \emph{University of California San Diego}\\
  \emph{kcyuen@eng.ucsd.edu, mtrivedi@eng.ucsd.edu}\\
}

\maketitle

\begin{abstract}
A key step to driver safety is to observe the driver's activities with the face being a key step in this process to extracting information such as head pose, blink rate, yawns, talking to passenger which can then help derive higher level information such as distraction, drowsiness, intent, and where they are looking. In the context of driving safety, it is important for the system perform robust estimation under harsh lighting and occlusion but also be able to detect when the occlusion occurs so that information predicted from occluded parts of the face can be taken into account properly. This paper introduces the Occluded Stacked Hourglass, based on the work of original Stacked Hourglass network for body pose joint estimation, which is retrained to process a detected face window and output 68 occlusion heat maps, each corresponding to a facial landmark. Landmark location, occlusion levels and a refined face detection score, to reject false positives, are extracted from these heat maps. Using the facial landmark locations, features such as head pose and eye/mouth openness can be extracted to derive driver attention and activity. The system is evaluated for face detection, head pose, and occlusion estimation on various datasets in the wild, both quantitatively and qualitatively, and shows state-of-the-art results.
\end{abstract}


% \fi
\IEEEpeerreviewmaketitle

\section{Introduction}

It was found that the most common factors of accidents are caused by distraction, drowsiness, and inattention \cite{klauer2010analysis}. At minimum in every year 2011 through 2014, a total of 29,500 fatal crashes involving 43,500 drivers resulted in 32,000 deaths where approximately 10\% of these crashes involved distraction. As a result of distraction, at least 380,000 are injured in crashes involving distracted drivers \cite{NHTSA_DD_2011}\cite{NHTSA_DD_2012}\cite{NHTSA_DD_2013}\cite{NHTSA_DD_2014}. Over a 5-year period from 2005 through 2009, there were at least 800 fatalities every year involving drowsy driving \cite{NHTSA_DrowsyD_2010}.

\begin{figure}
  \centering
  \begin{tabular}{c}
        \includegraphics[width=0.47\textwidth]{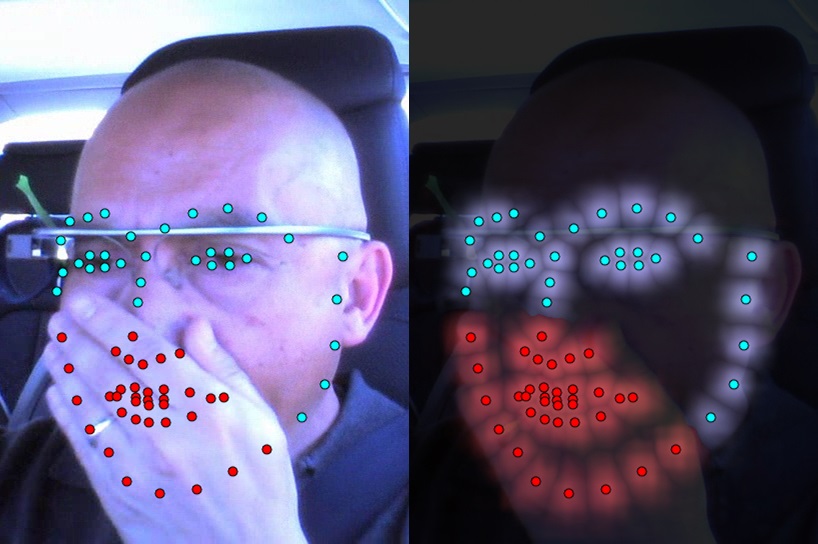}
  \end{tabular}
  \caption{(Left) Our proposed 68 facial landmark localization and occlusion estimation using the Occluded Stacked Hourglass showing non-occluded (blue) and occluded (red) landmarks. (Right) A visualization of the 68 heat maps output from the Network overlaid on the original image.}
  \vspace{-5mm}
  \label{fig:intro_fig}
\end{figure}

\begin{table*}[]
\centering
\caption{Comparison of Selected Studies on Face Analysis Methodologies}
\label{related-works-table}
\begin{tabular}{| L{3cm} | L{6cm} | L{3cm} | L{1.5cm} | L{1.8cm} |}
\hline
{\textbf{Research Study}} & {\textbf{Objective}} & {\textbf{Method}} & {\textbf{Number of Landmarks}} & \T{\textbf{Occlusion Estimation}}\B\\ \hline
\hline
{Viola, Paul, and Michael J. Jones, 2004 \cite{ViolaJones_IJCV2004}}                             & \T{Real-time face detection using computationally efficient features and a multi-stage classifier to allowing it to quickly reject background samples while focusing on the face-like windows in the later stages.}\B & {Cascade AdaBoost with Integral Image and Haar-like features} & {N/A} & {N/A}\\ \hline
{Zhu, Xiangxin, and Deva Ramanan, 2012 \cite{ZhuCVPR2012}}                               & \T{Face detection, discrete pose estimation, and landmark estimation by modeling facial landmarks as parts and using global mixtures to handle structural changes caused by different poses.}\B                       & {Deformable Part Model} &{38-68 (depending on pose)} & {N/A}                                                                   \T\B\\ \hline
{Yang, Shuo, et al., 2015 \cite{YangICCV2015}}                             & \T{Face detection from facial part responses (hair, eye, nose, mouth, beard).}                                                                                                                                      & {[Faceness] Convolutional Neural Network} & {N/A} & {N/A}\T\B\\ \hline
{Huang, Lichao, et al., 2015 \cite{huang2015densebox}}                                         & \T{Detects multiple different objects (faces and vehicles) and improves,detection accuracy when incorporating landmark localization during learning.}\B                                                               & {[DenseBox] Convolutional Neural Network} & {72} & {N/A}                                                                      \T\B\\ \hline
{Li, Lin, et al., 2015 \cite{li2015convolutional}}                                         & \T{Detects faces under large visual variations due to pose, expression and lighting using a CNN cascade to reject background regions while operating in low-resolution space to speed up performance, and evaluate higher probability candidates in the later stages at a higher resolution.}\B                                                               & {Convolutional Neural Network Cascade} & {N/A} & {N/A}                                                                       \T\B\\ \hline
\T{Farfade, Sachin Sudhakar, Mohammad J. Saberian, and Li-Jia Li, 2015 \cite{farfade2015multi}}\B & {Face detection under different angles and able to handle some occlusion by fine-tuning AlexNet}                                                                                                                  & {[DDFD] Convolutional Neural Network}  & {N/A} & {N/A} \T\B\\ \hline
\T{Burgos-Artizzu, Xavier P., Pietro Perona, and Piotr Dollár, 2013 \cite{burgos2013robust}}\B & {Performs landmark localization robustly under occlusion while also estimating occlusion of landmarks.}                                                                                                                  & {[RCPR] Cascaded Pose Regression}  & {29} & {80/40\% precision/recall} \T\B\\ \hline
\T{Ghiasi, Golnaz, and Charless C. Fowlkes, 2014 \cite{ghiasi2014occlusion}}\B & {Face detection, landmark estimation, and occlusion estimation using a hierarchical deformable part model, allowing the use of augmented positive samples using synthetic occlusion.}                                                                                                                  & {Hierachical Deformable Part Model}  & {68} & {81/37\% precision/recall} \T\B\\ \hline
\T{Yuen, Martin, Trivedi, 2016 \cite{yuen2016cnnfd}}\B & {Robust face detection under harsh lighting and occlusion, exploring the advantages and disadvantages of training with augmented samples.}                                                                                                                  & {Convolutional Neural Network}  & {N/A} & {N/A} \T\B\\ \hline
\T{Yuen, Martin, Trivedi, 2016 \cite{yuen2016cnnfdlme}}\B & {Robust face detection, landmark estimation, and head pose estimation under harsh lighting and occlusion using the AlexNet and Stacked Hourglass networks trained using augmented positive samples with synthetic occlusion.}                                                                                                                  & {Convolutional Neural Network}  & {68} & {N/A} \T\B\\ \hline

\hline
{This Work}                                                           & \T{Facial Landmark Localization, Occlusion Estimation, Head Pose Estimation and Face Detection Refinement using an Occluded Stacked Hourglass network and trained using augmented samples with synthetic occlusion.}\B                                                                                                     & {[Occluded Stacked Hourglass] Convolutional Neural Network} & {68} & {89/46\% precision/recall}                                                                       \T\B\\ \hline
\end{tabular}
\end{table*}
One way to minimize these events is by introducing an active safety system to assist the driver by monitoring their face in order to give alerts when the driver is not paying attention to a particular surrounding area. By tracking landmarks through the face along with eye information, head pose can be computed to provide the driver's facing direction. Indicators from eyes and mouth can provide information on eye closure duration, blink rates, or yawning to provide information on the driver's drowsiness or fatigue \cite{pei2002perclos}. Often when people are drowsy, they may cover their mouths when yawning or rub their eyes which both cause occlusion, which must be detected in order to avoid computing inaccurate measurements on driver's drowsiness or even the head pose. This allows for the system to warn the driver ahead of time of any unseen danger that they may not have seen or noticed and minimize the probability of an accident. By creating such a system also paves way to other systems which coordinates heads, eyes, and hands \cite{ohn2014head}, systems which looks at humans inside and outside vehicle cabins \cite{ohn2016looking}, or distraction \cite{hirayama2016classification}.

In an effort to create this system, the Stacked Hourglass, an existing deep neural network implementation designed for body pose joint estimation, is modified to take in a detected face window and output 68 occlusion heatmap images of facial landmarks which can be used to estimate landmark location and occlusion. The heatmap scores, which include rich landmark feature information of faces, also allows for the refinement of the original face detection input to the system, allowing the system to robustly reject any false positives that were part of the initial set of detections. The main contributions of this paper are: 1) heatmap-based facial landmark localization and occlusion estimation, 2) quantitative evaluation of face detection, head pose, and occlusion estimation, and 3) qualitative evaluation on various datasets and video sequences showing the robust generalization of trained model. The rest of the paper is as follows: the next section discusses related works in face detection, landmark localization, occlusion estimation, and the main contributions in this paper. Section III goes over the proposed system, discussing in detail about training sample generation, data augmentation, landmark localization, occlusion estimation, face detection refinement, and head pose estimation. Section IV evaluates the system on face detection and head pose estimation a challenging dataset found in both driving as well as real-world, presenting scenarios with harsh lighting and heavy occlusions on the face. Occlusion Estimation is evaluated on the Caltech Occluded Faces in the Wild, showing state-of-the-art results and improvements over previous work on occlusion. Concluding remarks and future direction is discussed in Section V.

%%%%%%%%%%%%%%%%%%%%%%%%%%%%%%%%%%%%%%%%%%%%%%%%%%%%%%%%%%%%%%%%%%%%%%%%%%%%%%%%
\section{Related Studies}

\begin{figure*}
  \centering
  \begin{tabular}{c}
        \includegraphics[width =0.95\textwidth]{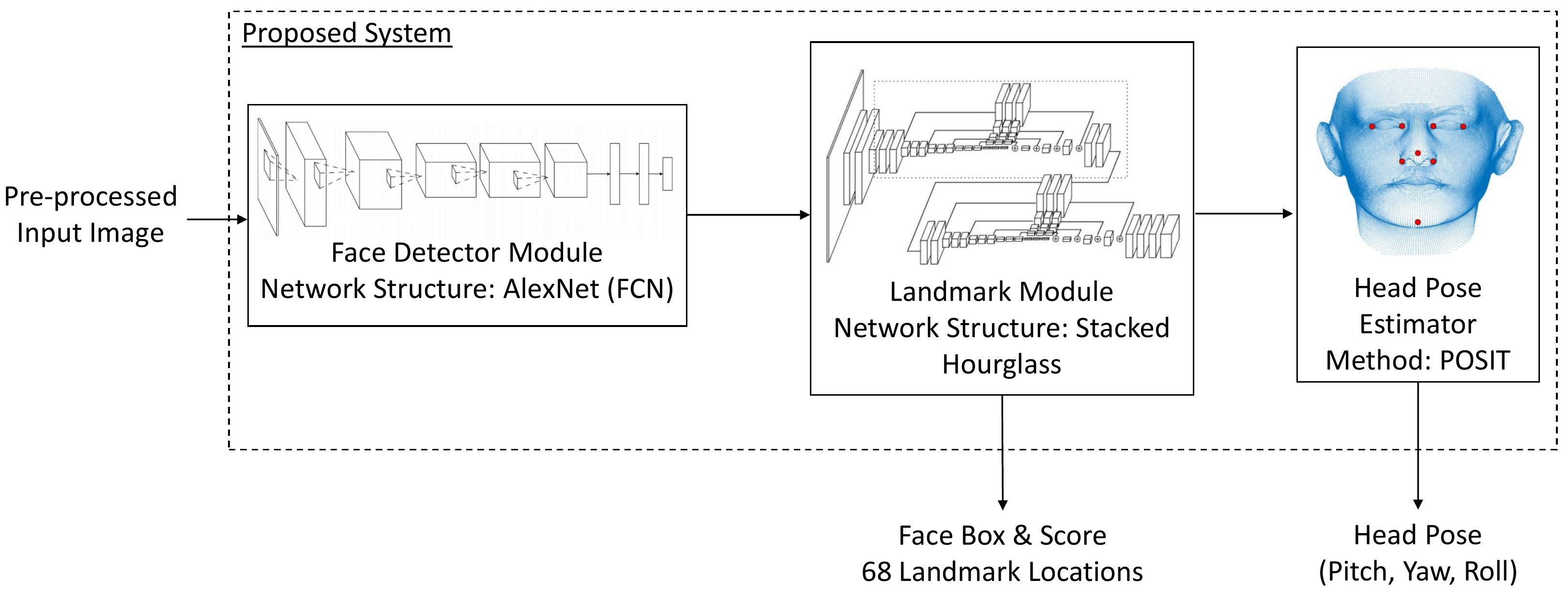}
  \end{tabular}
  \caption{Our proposed CNN pipeline system using two different CNN structures for detecting a face, localizing landmarks, estimating occlusion and head pose. Input pre-processing is done by CLA Histogram Equalization and passed through the face detector module. Resulting detected faces are processed by the landmark module to estimate landmark locations/occlusion, refine the original face detection output localization and score using landmark information. Head pose is estimated using the POSIT algorithm and a 3D generic face model with 8 selected rigid landmarks, indicated by the red dots on the model.}
  \vspace{-5mm}
  \label{fig:proposed_system}
\end{figure*}

In this section, several face detectors, facial landmark localizers, and occlusion estimation works are discussed along with their objective and methodologies. Early work in the face analysis domain include the Viola \& Jones face detector \cite{ViolaJones_IJCV2004} using a multi-stage cascade structure with computationally inexpensive features, and Zhu, et al. \cite{ZhuCVPR2012} detects faces, localizes up to 68 landmarks, and estimates discrete head pose using a deformable part model. Robust Cascaded Pose Regression \cite{burgos2013robust} localizes and estimates occlusion for 25 landmarks using robust shape-indexed features. Ghiasi, et. al. improves Zhu, et al.'s work on the DPM model by formulating a hierarchical deformable part model to improve landmark localization performance under occlusion while also estimating occlusion for each landmark.

More recent studies in the face domain are in deep learning which have achieved top performing results such as DDFD \cite{farfade2015multi}, Faceness \cite{YangICCV2015}, CascadeCNN \cite{li2015convolutional}, and DenseBox \cite{huang2015densebox}. DDFD fine-tunes AlexNet \cite{krizhevsky2012imagenet}, a well known CNN structure, to train a face detector by generating a score image and extracting detected faces. In Faceness, multiple DCNN components are trained to detect individual components of the face, e.g. eyes, nose, mouth, etc., and merges these individual responses to find a face. CascadeCNN achieves faster run times by cascading multiple CNNs to quickly reject non-face background areas in the early stages of the detector with less parameters, while the later stages have increasing complexity to reject more difficult false positives. A summary of related face detection papers is shown in Table \ref{related-works-table} which past work in face detection, their objective and methodolody, and whether or not they estimated landmarks and occlusion. A lot of past research has been done on face detection, some have researched landmark localization, and very few papers have researched into occlusion estimation. This work builds upon on previous two papers to further include occlusion estimation on top of face detection and landmark localization.

DenseBox introduced an end-to-end FCN framework directly predicting bounding box and scores, and showed that accuracy can be improved by incorporating landmark localization into the network. The Stacked Hourglass network \cite{newell2016stacked}, which is a very dense network used to predicting body joint locations given the location of a pedestrian, is the model our work is mainly based on. The network essentially pools to a low resolution, but with many features, then upsamples back up to a higher resolution to provide a heatmap output at the end of the network by fusing features from multiple resolutions, resulting in what visually looks like an hourglass. By placing hourglass modules together, it forms a stacked hourglass network.

Some advantages of a cascaded modular system the ability to take two existing networks that are known to work well as individual modules and use them together, and being able to choose smaller but quicker networks for the first stage to reject most of the false positives before entering larger but slower networks for more precise predictions. Control of precision and recall can be controlled at each stage of the outputs as well as being able to monitor where it went wrong in the middle of the system. Our work is mainly composed of two CNN systems, following the idea of Viola \& Jones and CascadeCNN by cascading classifiers together. The first stage is an AlexNet face detector, following closely to the work of DDFD, which performs face detection to find likely face window candidates. The Stacked Hourglass is used as the second stage, as Faceness and DenseBox found success in face detection by incorporating facial part responses, to perform a more expensive analysis on each of the candidate face windows and generating 68 facial landmark occlusion heatmap images. By developing a system that can estimate the location and occlusion of facial landmarks, an occlusion-aware face analysis system is possible.

%%%%%%%%%%%%%%%%%%%%%%%%%%%%%%%%%%%%%%%%%%%%%%%%%%%%%%%%%%%%%%%%%%%%%%%%%%%%%%%%
\section{Occluded Stacked Hourglass}

\begin{figure}
  \centering
  \includegraphics[width=1\linewidth]{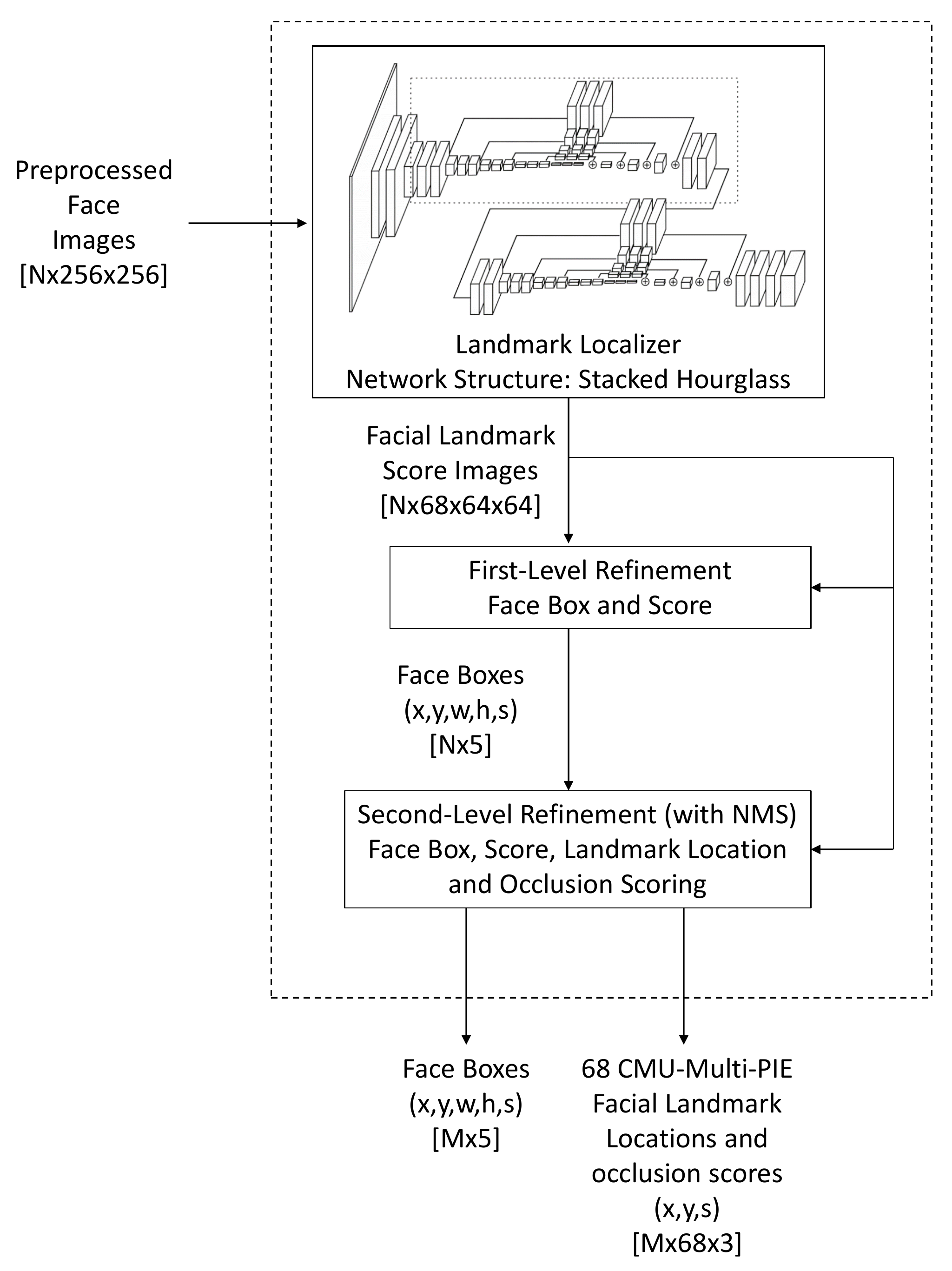}
  \caption{Our proposed landmark module network structure for estimating facial landmark and occlusion scores, and refinement of face boxes and scores to improve localization and precision. The system takes $N$ detected RGB face images which are preprocessed with CLA Histogram Equalization individually on each channel. These images are then fed into a Stacked Hourglass Network with 2 hourglasses. The resulting $N$ set of landmark score images of size $64\times64$, one for each of the $68$ landmarks, from the network are then post-processed using a 2-stage heuristic refinement method to extract $M$ set of face box and landmark locations, where $M$ is less than $N$ since the original input detections may be clustered together as the same face.}
  \vspace{-5mm}
  \label{fig:ch4_proposed_system}
\end{figure}

% Summary
This work builds on top of our previous work \cite{yuen2016cnnfd} which trains an AlexNet to detect faces while exploring the effects of augmentating training samples to mimic harsh lighting conditions and occlusion found while driving. In another of our previous work \cite{yuen2016cnnfdlme} shown in Fig. \ref{fig:proposed_system}, the Stacked Hourglass model was modified and trained to process detected faces and output 68 facial landmarks. To given an idea of the input-output functionality of the Stacked Hourglass, essentially it is designed to take an input image of fixed dimension and outputs multiple heatmap images, which are specified during training. The heatmap image training labels are generally designed such that a small Gaussian blob with amplitude 1 is placed at the location of the landmark relative to the input image with the rest of the heatmap being approximately zero. This results in heatmap image with values roughly in the range of 0 to 1, where high valued areas indicate a high confidence of the landmark being in that location.

In this paper, the Occluded Stacked Hourglass is presented, in which the training label heat map of the Stacked Hourglass is modified to also estimate occlusion for each facial landmark so that the system can tell which parts of the face is occluded, allowing future face analysis systems to be robust to any occlusions that may occur. The network is trained by following the original work of the Stacked Hourglass paper with a minor modification of including occlusion information in the heatmap ground truth labels. More specifically, if a landmark is labeled as occluded, then the Gaussian blob placed at the location of the landmark would have an amplitude of $-1$ instead of $+1$. The reasoning behind this is so that the location of the landmark can still be derived from looking at the magnitude of the heatmap, while the sign gives an indication of whether it's occluded or not. During training, techniques from our previous paper is used to apply augmentations to include artificial occlusion data. Heuristics are applied to the heatmaps resulting from the network output to extract facial landmarks and occlusion estimation. Head pose is estimated with POSIT algorithm \cite{dementhon1995model} with a 3D generic face \cite{martin2012design} and selected rigid landmarks. The initial bounding box proposal from the face detector is refined by generating a tighter-bounding box around the face using the location of estimated landmarks to provide better localization of the face, while the score is refined using occlusion confidence information. Our proposed modification to the landmark module is shown in Fig. \ref{fig:ch4_proposed_system}.

In the vehicle domain, the driver's face may undergo harsh lighting caused by the sun or occlusions due to objects such as sunglasses or hands. It is important that the system remains robust in these situations in order to continue monitoring the driver's face and behavior or send a signal when it is no longer able to monitor certain activities due to occlusion. Our contribution is two-fold: 1) the augmentation of training samples to include more harsh lighting and occlusion examples to improve the training data imbalance, and improve the robustness of the system under these situations, and 2) the use of the augmented occlusion to automatically create occlusion training labels to allow the landmark module to simultaneously localize landmarks and estimate occlusion.

 % Head pose is estimated with POSIT algorithm \cite{dementhon1995model} with a 3D generic face \cite{martin2012design} and selected rigid landmarks. 

% Dataset
% \subsection{Data Augmentation}

\subsection{Training Data Generation}
For landmark samples, a combined dataset of about 3,400 faces with 68 landmark annotations which follows the CMU Multi-PIE dataset format. The images are gathered from existing datasets: AFW \cite{zhu2012face}, HELEN \cite{le2012interactive}, LFPW \cite{belhumeur2013localizing}, and 300-W. The landmark annotations are provided by the 300-W challenge \cite{sagonas2016300}\cite{sagonas2013semi}\cite{sagonas2013300}. For negative samples (i.e. non-face), the 300-W challenge does not provide annotations for every face in the image, and so AFLW is used to generate negative samples where it is known that most of the faces have been annotated in this dataset. Face samples from AFLW was not used for landmark training since that dataset has only 21 landmarks annotated.

\textbf{Window Sampling}:
For positive face samples, only faces with a head pose roll angle of $\pm15^\circ$ is desired for the training set. To accomplish this, the face image is first de-rotated such that the face is no longer tilted (i.e. setting head pose roll angle to zero). This is done by rotating the image such that left-right symmetrical landmark pairs on the face, e.g. mouth corners, nose corners, eye corners, and brow corners, form a horizontal line as close as possible. Afterwards the face is randomly rotated by an angle chosen from a uniform distribution from -15 to 15 degrees. Since only landmarks are provided with the annotated face samples from the 300-W challenge dataset, an equivalent face box is computed by fitting a around the annotated landmarks, such that the box is as small as possible while still containing all landmarks. The ``annotated ground truth'' box is then generated by extending the height by 20\% from the top to include the forehead, as bounding box annotations from most face datasets include this portion of the face. A positive training sample of the face is then generated by randomly shifting and resizing a box such that it retains at least 70\% IOU (intersection of union) with the ground truth box. For Negative training samples are taken from the AFLW dataset if the square window is less than 5\% IOU. Here a low IOU threshold of 5\% is chosen since it is being used to train landmark localization, i.e. it is to minimize the amount of facial parts that appear in the negative training set, though higher values have not been experimented yet. About 90 positive samples per face in each image and 60 negative samples per image are extracted, resulting in about 217,000 face samples and 665,000 negative samples since there are more images in the AFLW dataset.

\textbf{Occlusions}:
The SUN2012 dataset, containing over 250,000 annotated regions and objects, is used artificially occlude the positive face samples for training by placing them at random locations of the face and at varying sizes. Since the distribution of object types was heavily biased towards region/objects such as wall, window, floor, etc., there is a danger of over-fitting the model to learn a particular object type from this dataset. To resolve this, an object type was first randomly selected then an object of that category type was randomly selected, both from a uniform distribution.

During initial training using only the SUN2012 dataset, there were issues with landmarks being marked as non-occluded even though they were covered by hands and/or sunglasses, and so a small dataset was collected composed of about 400 hands and 140 sunglasses with transparency masks already included from the various sources on the web. The object is chosen by first randomly selecting, from a uniform distribution, one of the three datasets, e.g. SUN2012, hands or sunglasses. An object is then randomly selected from within the chosen dataset, augmented, and placed on top of the positive sample. Due to resource constraints, all landmarks (regardless of whether they were occluded in the original image) are marked as non-occluded and only artificially occluded landmarks, using the method described here, are marked as occluded. It is expected that this should not harm the results significantly as most of the faces in the training dataset are non-occluded upon visual inspection, although the ideal situation would be to annotate the occluded landmarks in the original images for optimal results so that it is not trained with possible errors in ground truth.

\textbf{Histogram Equalization}: Since our domain resides in the driver cabin where harsh lighting may occur, improvements were found by applying contrast-limited adaptive histogram equalization (CLAHE) \cite{zuiderveld1994contrast} on the luma channel in the YCrCb space. All training and test images are preprocessed using CLAHE to ensure that the model is working with the same type of input in order to optimize performance.

\textbf{Flipping and Resizing Sampled Windows}: The samples are further augmented by randomly flipping left and right and are then resized to $256\times256$.

\subsection{Training Label Generation}
The hourglass network outputs $68$ $64\times64$ score images, which is smaller than the input image by a factor of 4. For positive samples, each score image is a 2-D Gaussian (see eq. \ref{eq:ch4_gauss_eq}) with mean $\mu$ being the scaled location (accounting for the factor of 4 scaling) of the landmark in image coordinates and standard deviation $\sigma=1.5$. The amplitude $A$ of the Gaussian is set to be $+1$ if the landmark is non-occluded, $-1$ if the landmark is occluded, and $0$ for all landmarks if it is a non-face negative sample. One motivation to setting occluded landmarks to have a value of -1 as opposed to some other value such as 0 or 0.5 is so that the magnitude of the score image looks the same regardless of whether the landmark is occluded or non-occluded, allowing us to robustly localize landmarks even under occlusion.

\begin{equation}\label{eq:ch4_gauss_eq}
G(x,y) = A \exp\left(- \left(\frac{(x-x_{LM})^2}{2\sigma_x^2} + \frac{(y-y_{LM})^2}{2\sigma_y^2} \right)\right)
\end{equation}

% \subsection{Model}

\subsection{Model Training}
The stacked 2-hourglass network is trained using published code provided by the Newell, Yang, \& Deng, 2016, original authors of the paper. Due to the large but powerful design, the network contains a convolutional layer with a stride of 2 and a max pooling layer at the beginning to downsample the input by a factor of 4. Each hourglass repeatedly pools the data to a smaller size then upsamples it back up in a symmetrical form. The hourglass takes advantage of the recently presented residual learning models \cite{he2015deep}, shown to make networks easier to optimize and improve accuracy with depth, by using it throughout the network. For more details on the hourglass network and residual modules, it is highly recommended to read through the original papers.

For training, the network uses 2 hourglass modules with 256 features per hourglass. On each batch iteration, 6 images are randomly chosen to be either a face or non-face with equal probability (i.e. a coin flip), and then randomly uniformly selects the image from the corresponding set (face or non-face) with replacement. The learn rate was manually adjusted based on training results and validation data: iteration 1-175K: $2.5\times10^{-4}$, iteration 175K-220K: $1.0\times10^{-4}$, iteration 220K-305K: $0.5\times10^{-4}$.

\subsection{Testing Phase}
All output boxes from the face detector are cropped from the image and fed through the hourglass network to output a set of 68 score images, one for each landmark. Predicted landmark locations are then extracted from these score images in order re-localize the detection box. A score for each landmark is calculated and summed to represent the refined score of the detection. In our initial experiments, different localized detections on the same face was found to agreed on the location of non-occluded landmarks, however occluded landmarks did not agree as much. There was slight noise in the occlusion estimation across different detections. To take care of these issues, noisy landmark localization and occlusion estimation are averaged out by summing the score images from overlapping detections, under the assumption that it is the same face. Landmark locations are then re-extracted from the normalized summed score images, and a score for each landmark is computed and thresholded to determine if a landmark is occluded. Each of these steps are explained in detail below.

\textbf{Initial Landmark Location}: Each detected face box will have 68 score images corresponding to each of the 68 landmarks, a visualization of this is shown in Fig. \ref{fig:intro_fig}. A similar procedure as in the face detector's testing phase with score images is used. The magnitude score image is thresholded at a value of 0.6 times the highest value in image, and connected component labeling is applied (e.g. MATLAB's bwlabel function) to group clusters of pixels together. The location of the landmark is then calculated as the weighted centroid of the group of pixels with the highest value in the magnitude score image (i.e. MATLAB's regionprops function). This is repeated for the remaining landmarks. The locations are then carefully transformed back from the $64\times64$ score image space to the resized $256\times256$ input face image space, and finally back to the original image coordinate space. The detection box is then re-localized based on the minimum and maximum of the x-y coordinates of all landmarks and extending the top of this box by 20\% as done during training sample generation of this chapter.

\textbf{Landmark Scores}: An ideal ground truth image is generated by placing a 2-D Gaussian at the predicted location of the landmark following the ground truth sample generation described earlier. The score of the $i^{th}$ landmark $S_{LM_i}$ is computed by computing the sum of squared magnitude difference of the pixels from its ideal ground truth image using eq. \ref{eq:ch4_lm_score}:

\begin{equation}\label{eq:ch4_lm_score}
S_{LM_i} = -\sum_{x=1}^{64}\sum_{y=1}^{64}(|I_{S_{LM_i}(ideal)}(x,y)| - |I_{S_{LM_i}(actual)}(x,y)|)^2
\end{equation}

where $|I_{S_{LM_i}(actual)}(x,y)|$ and $|I_{S_{LM_i}(ideal)}(x,y)|$ corresponds to a pixel at location (x,y) in the actual and ideal $64\times64$ magnitude score image, respectively, for a given landmark $i$. Using this scoring equation, an ideal landmark would have a score of 0, while a low scoring landmark would have a large negative value. By summing the individual landmark scores, the refined face detector score, $S_{RFD}$, is derived using eq. \ref{eq:ch4_fd_score}:

\begin{equation}\label{eq:ch4_fd_score}
S_{RFD} = \sum_{i=1}^{68} S_{LM_i}
\end{equation}

\textbf{Refining Landmark Location and Extracting Occlusion Score}: Looking at preliminary results, it was found that the 2-D Gaussian was not always negative as one would expect for occluded landmarks, making it difficult to classify a landmark as occluded. The location of these occluded landmarks were also jittery between different detections on the same face, compared to non-occluded landmarks where the locations generally agreed. To remedy this, results can be averaged together to produce a better prediction on occluded landmark location and scores by taking advantage of these observations.

Using the refined detection boxes and scores, detections are clustered together into groups by applying non-maximum suppression which creates a list by sorting the detections in descending order by score, selects the highest scoring detection in the list, groups all remaining detections in the list that has at least 20\% overlap (intersection over union) with the current detection and removes them from the list. Other overlap values have not been explored in detail in this work. This continues until all detections have been assigned its own groups.

Score images that belong to the same group are aligned by resizing, padding and translating from score image space into original image coordinate space using information from the original unrefined detection box, similar to the way the initial landmark locations are converted except this is working with score images rather than location points. The aligned score images are then summed together, and the process is repeated for all 68 landmarks resulting in 68 new score images.

Normalizing by the number of detections in the group was found to hurt the landmark score used to determine occlusion. This is because detections that were chosen as part of the group with very little overlap with the true face are likely to have a score image of a negative sample which is all zeros which would only lower the magnitude of the landmark score. Instead of normalizing by the number of detections in this group, the maximum absolute value across the 68 new score images is used for normalization so that the highest magnitude in any of the new score images is 1. The assumption is that at least one landmark is visible and is detected well by the network, and so the best scoring landmark is given a magnitude score of 1, and the rest of the landmarks are scored relatively to it. To extract the refined landmark locations, the same step as ``Initial Landmark Location'' in this subsection are applied using the new 68 landmark score images. The value of the pixel in the corresponding new score image at the refined landmark location is used as the landmark occlusion score.

\textbf{Head Pose Estimation from Landmarks}: Using a 8 rigid landmark locations selected from a 3D generic face model \cite{martin2012design} along with the corresponding estimated landmark positions, the head pose is estimated with the POSIT algorithm \cite{dementhon1995model}. The 8 rigid landmarks: two eye corners from each eye, nose tip, two nose corners, and the chin. Non-rigid landmarks such as mouth corners, which may move while the person is talking or yawning, are not used for pose estimation as they would deform from the 3D generic face model much more. Currently the use of occlusion estimation to disregard occluded landmarks from being used for head pose estimation has not been explored yet.

%%%%%%%%%%%%%%%%%%%%%%%%%%%%%%%%%%%%%%%%%%%%%%%%%%%%%%%%%%%%%%%%%%%%%%%%%%%%%%%%

\section{Evaluation}
This section will go over evaluation for each of the following: face detection, head pose, and occlusion estimation.

\subsection{Face Detection}
The evaluation for face detection refinement and head pose estimation is performed on the VIVA-Face dataset \cite{martin2016vision}, a test dataset challenge featuring a large range of lighting conditions and occlusions due to various objects and hand activities. The dataset contains a total of 607 faces from a total of 458 images selected for their challenging situation from 39 day-time naturalistic driving data video sequences, coming from data recorded with LISA-UCSD vehicles as well as YouTube videos (naturalistic driving to the best of our knowledge). 

\begin{figure}
  \begin{tabular}{ccc}
    \includegraphics[width=0.95\linewidth]{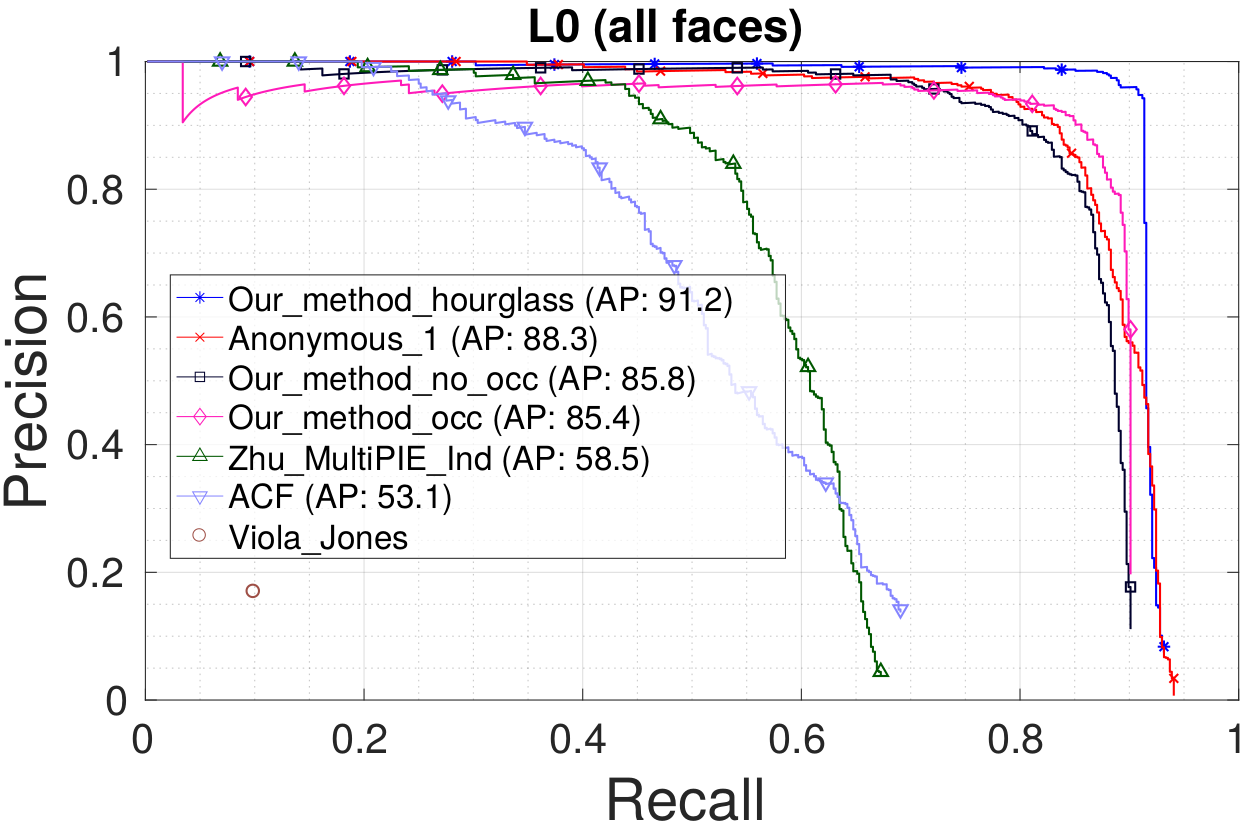}
  \end{tabular}
  \caption{Precision-Recall curve of the proposed method and baseline methods on the VIVA dataset containing 607 faces from 458 images.}
  \vspace{-3mm}
  \label{fig:PR_curves}
\end{figure}

Face detection is evaluated using the standard 50\% PASCAL overlap with precision-recall curve, shown in Fig. \ref{fig:PR_curves}. Methods such as Boosted cascade with Haar \cite{ViolaJones_IJCV2004}, Boosted cascade with ACF \cite{DollarPAMI2014} and Mixture of Tree Structure \cite{ZhuCVPR2012} are shown as a baseline for comparison. Two face detectors (Our\_method\_no\_occ and Our\_method\_occ) trained in our previous work is based on the AlexNet face detection module. For evaluating our new proposed system (Our\_method\_hourglass), improvement was found by concatenating the two outputs from both our previously trained face detectors, similarly done in ensemble methods, to improve recall rate. The curves show that the combination of the two face detectors has improved the recall, and with the second-stage refinement of face detection done by the landmark module, precision is significantly boosted particularly in the higher recall regions by a combination of improvement in face bounding box localization and false positive rejection using landmark data.

\subsection{Head Pose Estimation}
The head pose estimation estimation system is evaluated on the VIVA-Face Dataset. Yaw-angle evaluations of the head pose are shown and described in Table \ref{tbl:viva-face-yaw-table}, showing detection rate of the face, success rate of estimating the yaw within 15 degrees, mean and standard deviation of the absolute error. Our system is able to achieve a very high detection rate of at least 93.2\%, and estimate the yaw angle of the head pose with an average error of at least $5.6^\circ$ and a reasonable standard deviation of $10^\circ$, leading to a 93.4\% success rate of being able to estimate the head pose to within $15^\circ$ of the ground truth.

\begin{table}[ht]
  \begin{center}
    {\footnotesize
      \centering
      \begin{tabular}{|l|c|c|c|c|c|c|c|c|c|c|c|c|}
        \hline
        & \multicolumn{4}{c|}{L0 (all faces)}     \\ \hline
        Benchmark/Submission       & DR                       & SR                          & $\mu_{AE}$                       & $\sigma_{AE}$    \\ \hline
        Our\_method\_hourglass & \multicolumn{1}{l|}{$93.2\%$} & $93.4\%$ & $5.6^\circ$ & $10.0^\circ$  \\ \hline
        Anonymous\_4 & \multicolumn{1}{l|}{$60.8\%$} & $66.9\%$ & $13.9^\circ$ & $12.6^\circ$ \\ \hline
        Zhu\_MultiPIE\_Ind & \multicolumn{1}{l|}{$67.3\%$} & $63.1\%$ & $16.0^\circ$ & $16.5^\circ$ \\ \hline
      \end{tabular}
       \caption{Evaluation results from benchmark and submissions on yaw angle estimation of the head pose. DR, detection rate, is the percentage of images for which the face detector was able to detect a face with at least 50\% overlap with the ground truth face. SR, success rate, is the percentage of correctly detected faces for which the estimated yaw angle was within 15 degrees of the annotation. $\mu_{AE}$ and $\sigma_{AE}$ are the mean and standard deviation of the absolute yaw error (in degrees), respectively, calculated only from the correctly detected faces.}
      \label{tbl:viva-face-yaw-table}
    }
  \end{center}
  \vspace{-5mm}
\end{table}

\subsection{Occlusion Estimation}
\begin{figure}[h]
	% \vspace{-3mm}
  \centering
  \includegraphics[width=0.8\linewidth]{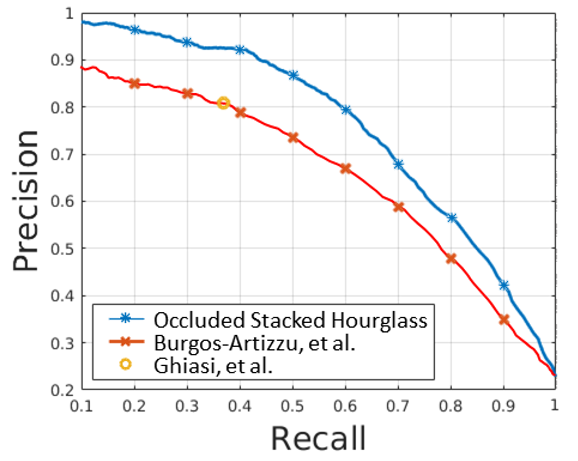}
  \caption{Occlusion detection precision/recall curves showing our trained Occluded Stacked Hourglass in comparison with other works.}
  % \vspace{-5mm}
  \label{fig:occ_eval}
\end{figure}

Occlusion estimation is evaluated using the COFW (Caltech Occluded Faces in the Wild) test dataset. The dataset contains faces with large variations in pose, expression, and an average occlusion of 23\%. Each face has 29 landmarks annotated indicating the location and whether or not each landmark is occluded. The evaluation results are  shown in Fig. \ref{fig:occ_eval}, showing significant improvement across all recall values in comparison to the robust cascaded pose regression by Burgos-Artizzu, et al. and hierarchical deformable part model by Ghiasi, et al. Ideally the occlusion score threshold should be around 0 (92/40\% precision/recall), since the ground truth was generated by setting the non-occluded and occluded landmark samples to have a score of 1 and -1, respectively, during the training process. However in practice, a low but positive score of a landmark also corresponds to the network not having high confidence in the detection and localization of the landmark which may be caused by some form of occlusion. It was found that a threshold of 0.2 (89/46\% precision/recall) was found to work better. resulting in higher recall at a small cost in precision.

\begin{figure}
  \centering
  \includegraphics[width=1\linewidth]{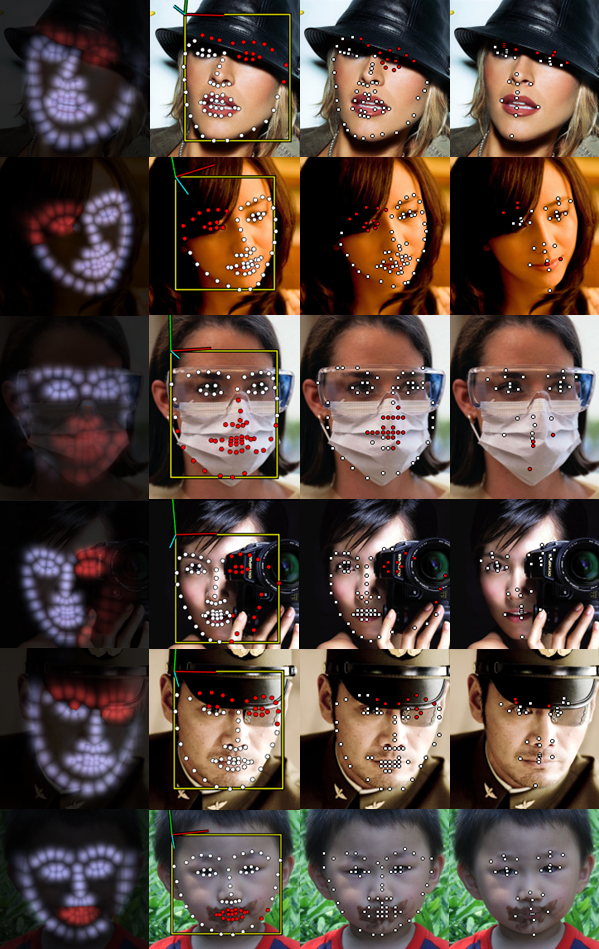}
  \caption{Example outputs from the Occluded Stacked Hourglass model applied on the COFW Dataset are shown in columns 1 \& 2. Column 1 shows the resulting 68 landmark heat maps (combined for visualization), showing non-occluded (blue) and occluded (red) landmarks. Column 2 shows the refined face detection box, estimated non-occluded landmarks (white) and occluded landmarks (red), and head pose. Columns 3 \& 4 show the results of Ghiasi, et al. (68 landmarks) and Burgos-Artizzu, et al. (29 landmarks), respectively, using their publicly available code.}
  \vspace{-5mm}
  \label{fig:ch4_example_outputs_cofw}
\end{figure}

\begin{figure*}[]
%\todo{crop images nicely}
  \centering
  % \begin{tabular}{p{8.7cm}}
    % \multicolumn{1}{c}{
      % Column 1
      \begin{subfigure}[b]{0.3180\textwidth}
        \includegraphics[width =\textwidth]{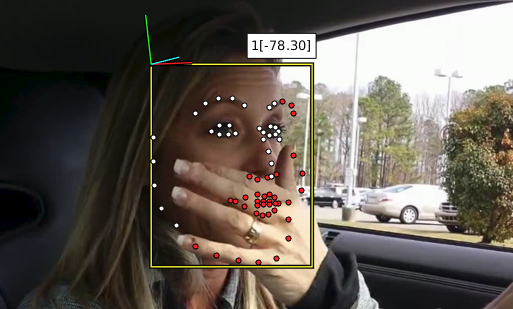}\label{}
        \vspace{-1.1\baselineskip}
        \includegraphics[width =\textwidth]{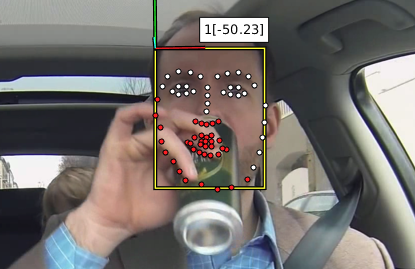}\label{}
      \end{subfigure}%
      \hspace{-.17cm}
      % Column 3
      \begin{subfigure}[b]{0.2960\textwidth}
        \includegraphics[width =\textwidth]{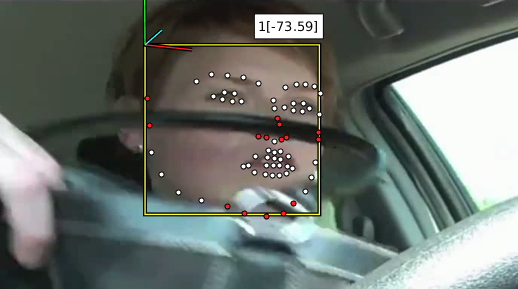}\label{}
        \vspace{-1.1\baselineskip}
        \includegraphics[width =\textwidth]{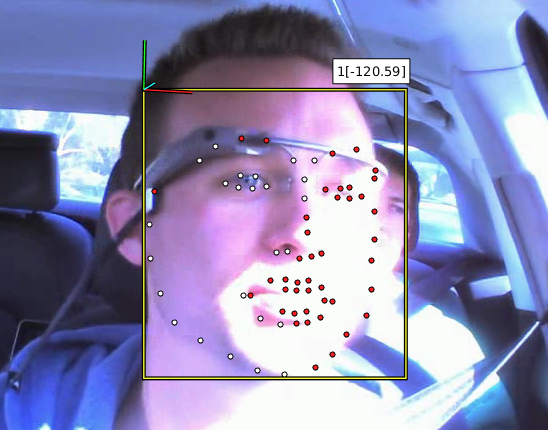}\label{}
      \end{subfigure}%
      \hspace{-.17cm}
      \begin{subfigure}[b]{0.3010\textwidth}
        \includegraphics[width =\textwidth]{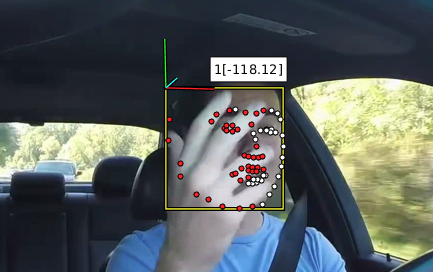}\label{}
        \vspace{-1.1\baselineskip}
        \includegraphics[width =\textwidth]{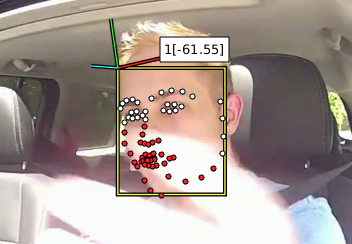}\label{}
      \end{subfigure}%
      \hspace{-.17cm}
      % Column 1
      \begin{subfigure}[b]{0.29671\textwidth}
        \includegraphics[width =\textwidth]{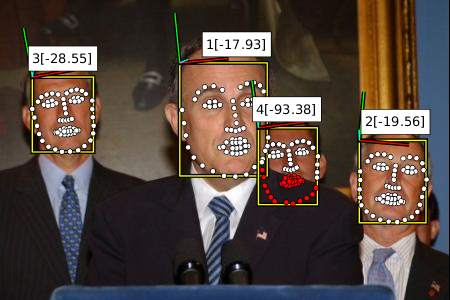}\label{}
        \vspace{-1.1\baselineskip}
        \includegraphics[width =\textwidth]{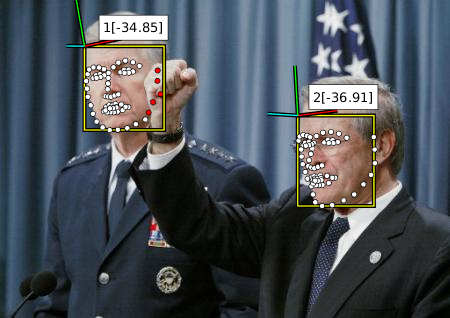}\label{}
        \vspace{-1.1\baselineskip}
        \includegraphics[width =\textwidth]{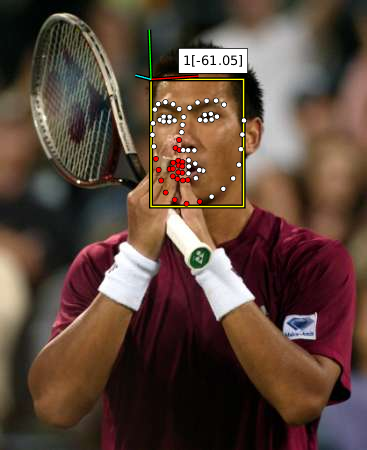}\label{}
      \end{subfigure}%
      \hspace{-.17cm}
      % Column 2
      \begin{subfigure}[b]{0.35\textwidth}
      \includegraphics[width =\textwidth]{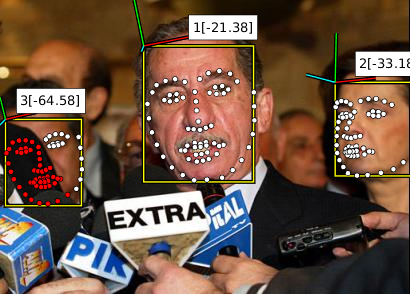}\label{}
        \vspace{-1.1\baselineskip}
        \includegraphics[width =\textwidth]{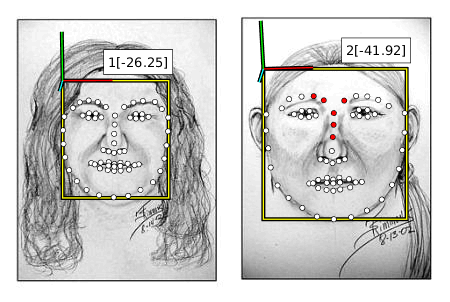}\label{}
        \vspace{-1.1\baselineskip}
        \includegraphics[width =\textwidth]{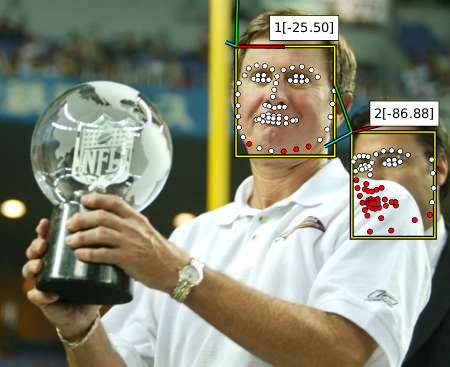}\label{}
      \end{subfigure}%
      \hspace{-.17cm}
      % Column 3
      \begin{subfigure}[b]{0.2696\textwidth}
      \includegraphics[width =\textwidth]{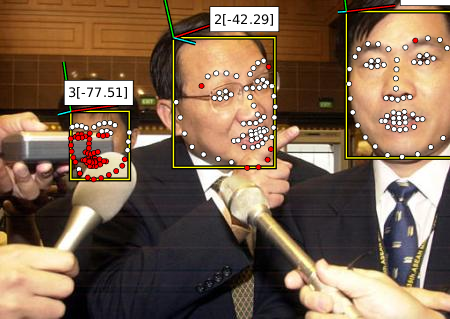}\label{}
        \vspace{-1.1\baselineskip}
        \includegraphics[width =\textwidth]{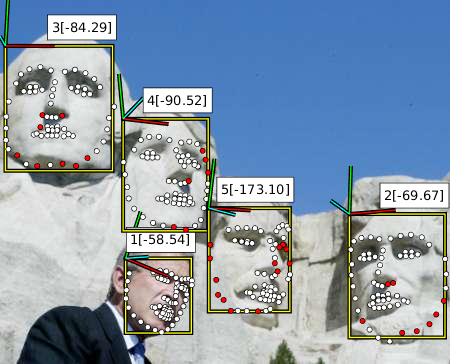}\label{}
        \vspace{-1.1\baselineskip}
        \includegraphics[width =\textwidth]{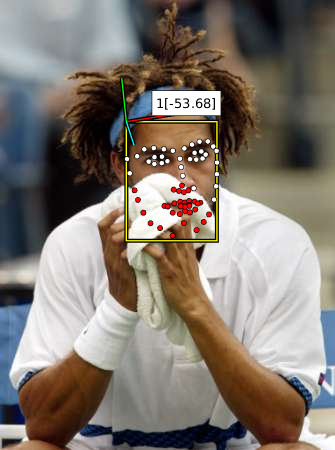}\label{}
      \end{subfigure}%
      \hspace{-.17cm}
    % }
  % \end{tabular}
  \vspace{5mm}
  \caption{Example outputs from our Occluded Stacked Hourglass model applied on the VIVA-Face Dataset (first two rows) and FDDB (bottom three rows), showing the refined face detection box, estimated non-occluded landmarks (white) and occluded landmarks (red), and head pose.}
  \label{fig:ch4_example_outputs_viva}
\end{figure*}

A comparison on a few images from the COFW dataset of our method with Ghiasi, et al. and Burgos-Artizzu, et al. are shown in Figure \ref{fig:ch4_example_outputs_cofw}. Our method shows to out-perform the other two methods in on most parts in both landmark localization and occlusion estimation. Landmark localization can still be improved, particular in areas where there is heavy occlusion areas such as the images on rows 3 \& 4 where a mask or large camera is blocking a large percentage of the face. Fig. \ref{fig:ch4_example_outputs_viva} shows that the system is robust in other datasets such as VIVA-Face and FDDB, while also proving to be generalizable to even statues or drawings of faces. The system is shown to be able to handle and predict occlusions caused by hands, shoulders, various objects, as well as harsh sun light that washes out the features of the face to the point that it would be considered occluded.

\begin{figure*}
  \centering
  \begin{tabular}{c}
        \includegraphics[width =1\textwidth]{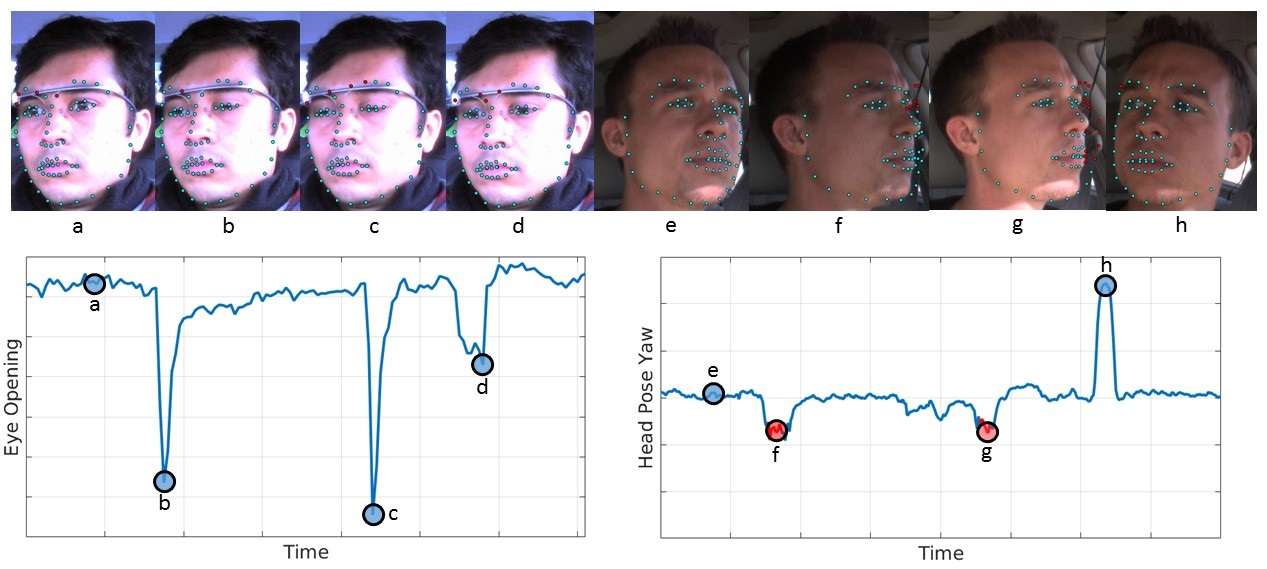}
  \end{tabular}
  \caption{The proposed system applied on videos of the driver's face. Left sequence shows analysis on driver's landmarks around the eyes which computes the eye opening: the driver's eye opened (a), blinking (b,c), and looking at the speedometer (d). Right sequence shows the driver's head pose yaw angle: the driver's head facing straight ahead (a), looking left (b,c), and looking right (d), where red indicates the scores for landmarks used in the head pose estimation did not meet a required threshold.}
  \vspace{-5mm}
  \label{fig:blink_and_headpose_timeline}
\end{figure*}

\begin{figure*}
  \centering
  \begin{tabular}{c}
        \includegraphics[width =1\textwidth]{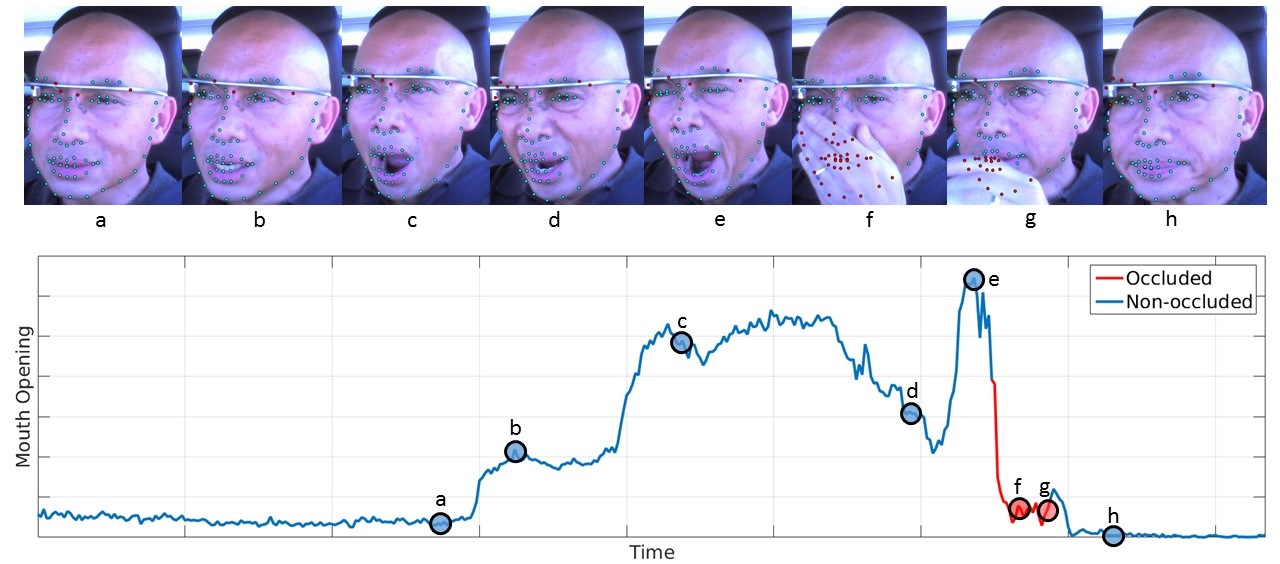}
  \end{tabular}
 	\caption{Example application on another video sequence where the driver is yawning. The mouth opening graph is computed from the area enclosed by the six inner mouth landmarks. The driver's mouth are shown closed at (a), then slowly opens from (b) to (e). (f) and (g) shows red occluded landmarks caused by a hand covering the mouth. The red line on the graph indicates that the mouth is occluded and is computed based on a certain threshold of landmarks estimated as occluded in the area of the mouth. Finally (h) shows driver's mouth no longer being occluded and closed. A video demonstration of the Occluded Stacked Hourglass is shown here: \href{http://cvrr.ucsd.edu/kcyuen/occluded_stacked_hourglass/}{http://cvrr.ucsd.edu/kcyuen/occluded\_stacked\_hourglass/}}
  \vspace{-5mm}
  \label{fig:yawn_timeline}
\end{figure*}

The system currently does not implement any background subtraction or tracking over time at any stage as it is originally designed to work on single images at a time, particularly the VIVA-Face dataset which contains images of drivers from many different camera views with no temporal relation between images. However, results on image frames from a video clips processed individually are shown in Fig. \ref{fig:blink_and_headpose_timeline} showing the use of eye landmarks to compute the eye opening which can be used for detecting blinks or when the driver is looking down. In the same figure, the head pose yaw angle provides where the driver is facing towards. Fig. \ref{fig:yawn_timeline} shows the use of mouth landmarks to compute the mouth opening to detect when the driver yawns. The red portion shows occlusion when the driver naturally covers their mouth with their hand while yawning. Note that the y-axis values have been normalized based on the minimum and maximum observed values of the drives. When deploying this as a video analysis tool, background subtraction and tracking at the face detection, landmark, and head pose stages should be considered in order to reduce false positives and increase robustness of the system by exploiting temporal consistency.

%%%%%%%%%%%%%%%%%%%%%%%%%%%%%%%%%%%%%%%%%%%%%%%%%%%%%%%%%%%%%%%%%%%%%%%%%%%%%%%%

\section{Concluding Remarks}
In this paper, a framework has been proposed to detect a face with high recall rates using the AlexNet, then processing the detected faces for estimating landmark location and occlusion using the Stacked Hourglass Network, then subsequently re-estimating a score for the detection for a high precision rate, and finally using the landmarks to estimate a head pose. It is shown that this results in a high precision and recall rate for face detection along with a very high success rate for estimating head pose. This paper also introduced the Occluded Stacked Hourglass, a method to train the Stacked Hourglass model with occlusion information embedded into the heat maps labels, we were able to successfully outperform previous state-of-the-art methods on estimating facial landmark occlusion information. Using the information on where the occlusion occurred on a driver's face allows future head, eye, or mouth algorithms to become more robust in the case that any objects or sunlight occlude any related facial parts.

A current issue in the Stacked Hourglass module is that it was not trained to handle profile faces. When the head is turned away from the camera with only half of the face visible (self-occlusion), the landmark module will have issues estimating the location of the missing landmark. This is due to the 68-landmark training data used which is composed of faces where all 68 landmarks are not occluded by self-occlusion. This issue is planned to be resolved in future work by training a separate model for profile-view with fewer points such as 38-profile/68-frontal landmarks as annotated in the CMU-MultiPIE dataset or the newly released Menpo Dataset \cite{zafeirioumenpo}. The introduced system allows for future research and development of future higher level facial analysis of driver's state and focus such as distraction, drowsiness, emotions, gaze zone estimation \cite{vora2017generalizing}\cite{martin2017gaze}, and other social activities involving the face.

\section{Acknowledgments}
The authors would like to specially thank Dr. Sujitha Martin and reviewers for their helpful suggestions towards improving this work. The authors would also like to thank our sponsors and our colleagues at the Computer Vision and Robotics Research Laboratory (CVRR) for their assistance.

% or


% \appendices

% % you can choose not to have a title for an appendix
% % if you want by leaving the argument blank
% \section{}

% % use section* for acknowledgment
% \section*{Acknowledgment}


\ifCLASSOPTIONcaptionsoff
  \newpage
\fi



%\bibliographystyle{IEEEtran}


% \end{thebibliography}
\bibliographystyle{IEEEtran}
\bibliography{bare_jrnl_bib}


\begin{IEEEbiography}[{\includegraphics[width=1in,height=1.25in,clip,keepaspectratio]{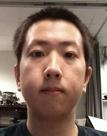}}]{Kevan Yuen} received the B.S. and M.S. degrees in electrical and computer engineering from the University of California, San Diego, La Jolla. During his graduate studies, he was with the Computer Vision and Robotics Research Laboratory, University of California, San Diego. He is currently pursuing a PhD in the field of advanced driver assistance systems with deep learning, in the Laboratory of Intelligent and Safe Automobiles at UCSD.
\end{IEEEbiography}

\begin{IEEEbiography}[{\includegraphics[width=1in,height=1.25in,clip,keepaspectratio]{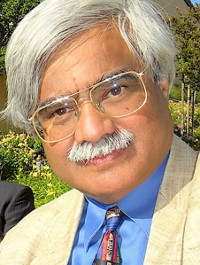}}]{Mohan Manubhai Trivedi} is a Distinguished Professor of and the founding director of the UCSD LISA: Laboratory for Intelligent and Safe Automobiles, winner of the IEEE ITSS Lead Institution Award (2015). Currently, Trivedi and his team are pursuing research in distributed video arrays, active vision, human body modeling and activity analysis, intelligent driver assistance and active safety systems for automobiles. Trivedi’s team has played key roles in several major research initiatives. These include developing an autonomous robotic team for Shinkansen tracks, a human-centered vehicle collision avoidance system, vision based passenger protection system for “smart” airbag deployment and lane/turn/merge intent prediction modules for advanced driver assistance. Some of the professional awards received by him include the IEEE ITS Society’s highest honor “Outstanding Research Award” in 2013, Pioneer Award (Technical Activities) and Meritorious Service Award by the IEEE Computer Society, and Distinguished Alumni Award by the Utah State University. Three of his students were awarded “Best Dissertation Awards” by professional societies and a number of his papers have won “Best” or “Honorable Mention” awards at international conferences. Trivedi is a Fellow of the IEEE, IAPR and SPIE. Trivedi regularly serves as a consultant to industry and government agencies in the U.S., Europe, and Asia.
\end{IEEEbiography}

% \begin{IEEEbiographynophoto}{Kevan Yuen}

% \begin{IEEEbiographynophoto}{Sujitha Martin}

% \begin{IEEEbiographynophoto}{Mohan Trivedi}

%\newpage



%\vfill


\end{document}